\documentclass[preprint,5p,times,twocolumn]{elsarticle}

\usepackage{amssymb}
\usepackage{amsmath}

\usepackage{url}
\usepackage{hyperref}
\usepackage{pifont}
\usepackage{multirow}
\usepackage{multicol}
\usepackage{color}
\usepackage{xcolor, soul}
\usepackage{colortbl}

\soulregister{\hl}{1}
\newcommand{\highlighttext}[3][RGB]{
    \begingroup
    \definecolor{hlcolor}{#1}{#2}\sethlcolor{hlcolor}
    \hl{#3}
    \endgroup
}

\definecolor{myyellow}{RGB}{253,233,149}
\definecolor{mycyan}{RGB}{167,224,246}
\definecolor{mygreen}{RGB}{166,225,197}
\definecolor{myred}{RGB}{255,159,174}
\definecolor{mymor}{RGB}{225,167,251}
\definecolor{myorange}{RGB}{255,159,127}

\makeatletter
\def\ps@pprintTitle{%
    \let\@oddhead\@empty
    \let\@evenhead\@empty
    \let\@oddfoot\@empty
    \let\@evenfoot\@empty}
\makeatother

\journal{}

\begin{document}

\begin{frontmatter}


\title{Unraveling the Capabilities of Language Models in News Summarization}

\author[tau]{Abdurrahman Odabaşı} 
\author[ytu,tau]{Göksel Biricik} 

\affiliation[tau]{organization={Department of Computer Engineering, Turkish-German University},
            postcode={34820}, 
            state={Istanbul},
            country={Türkiye}}

\affiliation[ytu]{organization={Department of Computer Engineering, Yıldız Technical University},
            postcode={34220}, 
            state={Istanbul},
            country={Türkiye}}

\begin{abstract}

Given the recent introduction of multiple language models and the ongoing demand for improved Natural Language Processing tasks, particularly summarization, this work provides a comprehensive benchmarking of 20 recent language models, focusing on smaller ones for the news summarization task. In this work, we systematically test the capabilities and effectiveness of these models in summarizing news article texts which are written in different styles and presented in three distinct datasets. Specifically, we focus in this study on zero-shot and few-shot learning settings and we apply a robust evaluation methodology that combines different evaluation concepts including automatic metrics, human evaluation, and LLM-as-a-judge. Interestingly, including demonstration examples in the few-shot learning setting did not enhance models’ performance and, in some cases, even led to worse quality of the generated summaries. This issue arises mainly due to the poor quality of the gold summaries that have been used as reference summaries, which negatively impacts the models’ performance. Furthermore, our study’s results highlight the exceptional performance of GPT-3.5-Turbo and GPT-4, which generally dominate due to their advanced capabilities. However, among the public models evaluated, certain models such as Qwen1.5-7B, SOLAR-10.7B-Instruct-v1.0, Meta-Llama-3-8B and Zephyr-7B-Beta demonstrated promising results. These models showed significant potential, positioning them as competitive alternatives to large models for the task of news summarization.
\end{abstract}


\end{frontmatter}


\section{Introduction}\label{sec:intro}

In today’s digital age, the amount of data being produced has grown exponentially.
The rapid increase in data, particularly in the news sector, has made it crucial to summarize information quickly and accurately to stay informed. 
News plays an integral role in our daily lives by keeping us updated about global events and shaping our perspectives, knowledge, and opinions. 
However, staying well-informed without feeling overwhelmed is challenging \cite{elkassas20, mridha21}. 
This study, therefore, focuses on the task of News Summarization, which involves presenting the key facts and significant details from a news article in a clear and concise format, allowing individuals to stay effectively informed about current events. 
Manual summarization, while good at maintaining the original meaning of the text, is impractical due to its time-consuming nature. But one valuable solution to this problem is Automatic Text Summarization (ATS), which efficiently aims to condense lengthy articles into brief summaries, focusing exclusively on the main aspects of the original content, saving time, effort and resources by making it simpler to rapidly comprehend the primary concepts without reading the entire document \cite{elkassas20, zhang22, hew22, yadav22}.

As Generative Artificial Intelligence (GenAI) technologies continue to advance, along with the increasing number of Language Models (LMs) introduced every day, there is growing interest in leveraging their capabilities to enhance the efficiency and accuracy of not only news summarization but various Natural Language Processing (NLP) tasks \cite{cao23, zhao23}.
The release and widespread use of LMs like ChatGPT have undoubtedly not only showcased AI’s potential but have also increased public awareness of its capabilities \cite{zhao23, hadi23, kalyan24}. 
Imagine a world where you could instantly grasp the key points of any lengthy news article with just a brief glance---this is what we aim to achieve with LMs. 

While Large Language Models (LLMs) like GPT-4, Claude, etc. have demonstrated impressive capabilities, their substantial size—with parameters ranging from tens to hundreds of billions or maybe even more—leads to significant challenges including high computational power requirements, increased latency, costly training and maintenance, and limited flexibility. 
This has driven researchers to explore smaller language models. 
These compact models present a promising alternative for LLMs since they are more computationally efficient, require less memory and storage space, and offer more cost-effective deployment options, making them particularly interesting candidates for investigating effective NLP solutions. 
Nevertheless, important questions remain: Can such smaller language models manage the information load and ensure that critical news reaches its audience both efficiently and effectively through summarization? How effectively can these smaller-scale language models handle news summarization tasks while balancing efficiency and performance?  How well do different LMs perform in summarizing news? Which are good, which are better, and which should we avoid?

There are a few studies in the field of news summarization where a small number of LMs of different sizes were evaluated, but these models are becoming outdated and limited compared to recent ones \cite{goyal22, zhang23, basyal23}.
This study attempts to fill this knowledge gap and address previous questions by evaluating the performance of various small and mid-sized LMs in the context of news summarization and comparing their performance with that of larger models, focusing on both zero- and few-shot learning scenarios.
Through systematic analysis, this study seeks to identify these models’ strengths and drawbacks, ultimately determining the most effective LMs in summarizing news articles.
This will contribute to the development of more sophisticated and reliable AI-powered news summarization systems, ensuring people stay well-informed of important news content.

This work presents several original contributions to the fields of GenAI and NLP, specifically in the context of news summarization task.
We conclude our research’s major contributions as follows:

\begin{itemize}

\item We conducted a comprehensive benchmark of 20 contemporary language models’ performance in the news summarization task, considering zero-shot and few-shot in-context learning scenarios, offering new insights into the capabilities of the benchmarked models, as the large number of models evaluated simultaneously is significant compared with the number of models considered in prior works. 
This extensive benchmark delivers a broad comprehension of the capabilities and limitations of recent models.

\item We employed a multifaceted evaluation approach, which included automatic metrics, human evaluation, and AI-based evaluation. 
By utilizing these diverse methods, we ensured a more reliable and comprehensive analysis, offering a nuanced understanding of the level of quality of LM-generated summaries.

\end{itemize}

We hope our work highlights the strengths and weaknesses of various LMs for news summarization and guides future improvements in summarization tasks.

\section{Related Work}\label{sec:related_work}

Automatic text summarization is one of the significant tasks in NLP, with a rich research history. 
It began with Luhn \cite{luhn58}, who came up with the idea of summarizing scientific documents by extracting the most significant sentences. 
Early research on both abstractive and extractive summarization relied on various approaches along the course of the research journey, such as statistical, graph-based, structure-based, clustering-based, fuzzy logic-based, and machine learning approaches. 
Over time, deep learning techniques, including Feed Forward and Recurrent Neural Networks (RNNs), advanced summarization capabilities, with Sequence-to-Sequence (Seq2Seq) RNNs becoming a notable standard \cite{elkassas20, mridha21}.

However, the introduction of attention mechanisms and the Transformer architecture by Vaswani et al. \cite{vaswani17} catalyzed advancements in GenAI we see nowadays and caused an uproar in the NLP community, as Transformer models surpassed previous methods across a wide range of NLP tasks, leading to numerous studies investigating earlier Transformer-based Encoder-Decoder Language models such as BART, Pegasus, and T5 on news summarization tasks (\cite{gupta21, karkera22} and others).

With Transformer-based language models becoming increasingly advanced, featuring decoder-only architecture, enhancing their own generative capabilities, and encompassing a broader understanding of language structure and knowledge across various fields more than ever before, it has become essential to benchmark these models on specific NLP tasks, such as news summarization, considering that numerous models are being released by both commercial companies and open-source communities across different sizes and parameter scales \cite{zhao23, minaee24}. 
Yet recent developments have shown promising results of LMs built on non-Transformer architectures \cite{gu23}, which also show significant promise. 
However, these non-Transformer architectures fall outside the scope of our current study, which focuses exclusively on recent Transformer-based models.

In 2022, Goyal et al. examined how well models of different types performed on the news summarization task. 
They assessed a general-purpose model, GPT-3, against a task-specific fine-tuned model, BRIO, and another model optimized for numerous tasks, T0. 
They extended the scope of generic summarization to include keyword-based summarization, specifically requesting a summary of the text with an emphasis on a certain keyword (topic, person, etc.). 
In all their experiments, it was consistently observed that GPT-3 received lower scores on automatic metrics compared to other models. However, it significantly surpassed them in terms of human evaluation \cite{goyal22}.

Zhang et al. conducted a thorough evaluation of LMs of different sizes. 
The number of benchmarked models has been expanded to ten, comparing different versions of OpenAI GPT-3 and InstructGPT models including Ada, Curie and Davinci versions, as well as other models like Anthropic-LM-v4, Cohere-XL, GLM and OPT. 
They performed both zero-shot, and few-shot prompting using five examples. 
Their research discovered that InstructGPT models –especially the davinci version– were capable of achieving news summarization levels that were comparable to those of human summaries \cite{zhang23}.

Another study was carried out by Basyal and Sanghvi later in 2023 using popular news summarization benchmarks to compare several newer models—more precisely, tuned versions of the falcon-7b-instruct, mpt-7b-instruct, and the first model behind ChatGPT, the text-davinci-003. 
According to their experiments, text-davinci-003 outperformed the others \cite{basyal23}.

To wrap up, while prior research has evaluated a limited number of LMs for news summarization, some of these models are outdated and do not reflect the capabilities of the latest advancements in the Generative AI field.

As previously stated, recent advancements are credited to the introduction of Transformer architecture, the availability of vast datasets, and improved computational resources.
However, we note that the number of model parameters has continuously increased, from hundreds of millions to tens of billions, even to hundreds of billions. 
This increase in parameters is correlated with enhanced language understanding and better performance on complex tasks. 
Thus, the ongoing trend of “larger is better” has led some to propose a new Moore’s law for LMs. 
Nevertheless, this is not always practical due to the significant costs and complexities involved in training these large models. 
While large models demonstrate emergent properties such as breaking down complex tasks, reasoning, and problem-solving, smaller models are still valuable, as they can be easily fine-tuned for specific applications like reading comprehension or summarization, achieving excellent results, considering also that the training approaches for such models are evolved and new special techniques are continuously being researched \cite{minaee24}. 
Thus, our research seeks to systematically assess the effectiveness of several recent small and mid-sized LMs in In-Context learning scenarios.

\section{Experimental setup}\label{sec:experimental_setup}

\subsection{Datasets}\label{subsec:datasets}

In order to be able to assess the performance of different language models on the task of news summarization, several benchmark datasets have been utilized. 
These benchmark datasets share certain attributes that align with our research: Firstly, our focus was primarily on English datasets, comprised of English news articles and their related summaries, due to the wide usage and availability of resources in English.
Secondly, datasets must be dedicated to the Single-Document News Summarization task (SDS), which involves extracting essential points from a singular news story and compressing them into a succinct summary. 
While Multi-Document Summarization (MDS) that combines and synthesizes information from several news articles, represents another important research direction, it falls outside of the scope of this paper. 
Therefore, we distinguish our datasets from others such as Document Understanding Conferences (DUC) benchmarks \cite{harman04} and Multi-News \cite{fabbri19}, which are intended for use in the Multi-Document News Summarization task. 
Thirdly, the gold summaries provided in the dataset should ideally be abstractive, human-generated, and reasonably condensed. 
Lastly, the datasets should be widely recognized, characterized by their large size, comprising hundreds of thousands of news articles, and publicly accessible. 
Three prominent benchmarks that meet all these criteria are CNN/Daily Mail, also known as (CNN/DM), Newsroom, and Extreme Summarization, also known as (XSum) (presented in Table \ref{tab:datasetsinfo}) \cite{cnndm1, cnndm2, newsroom, xsum}.

\begin{table}[t]
    \centering
    \scriptsize
    \begin{tabular}{ccccccccc}
        \hline
        Name &  Year & \#Sources & Train Set & Val Set & Test Set \\
        \hline
        CNN/DM \cite{cnndm1, cnndm2, cnndmhf} & 2015/2016 & 2 & 287,113 & 13,368 & 11,490 \\
        Newsroom \cite{newsroom} & 2018 & 38 & 995,041 & 108,837 & 108,862 \\
        XSum \cite{xsum, xsumhf} & 2018 & 1 & 204,045 & 11,332 & 11,334 \\
        \hline
    \end{tabular}
	\caption[General information about the popular datasets in the field of News Summarization.]{General information about the popular datasets in the field of News Summarization.}
	\label{tab:datasetsinfo}
\end{table}

Initially, the goal of the CNN/Daily Mail (CNN/DM) dataset was to facilitate the tasks of Passage-based Question Answering and Reading Comprehension, as it contained news articles as passages and abstractive short summaries in the form of bullet points \cite{cnndm1}. 
However, in 2016 with a simple straightforward modification, Nallapati et al. adapted the original dataset to serve as a benchmark for ATS task by concatenating the highlight bullets to form a single, multi-sentence summary for each news article, where each bullet \cite{cnndm2}. 
Although the method used to generate the gold summaries (highlights within the article) may not be optimal for evaluating ATS task and summaries of good quality should be in the format of a coherent paragraph, not separated sentences, where each of them may explain something different, this dataset still seems to be one of the famous benchmarks in this field \cite{chen16}.

In 2018, Grusky et al. released the Newsroom dataset, a large-scale collection sourced from 38 distinct major news publishers (such as Aljazeera, BBC, CNBC, Fox Sports, NY Daily News, Reuters, etc.), distinguishing itself from previous datasets that relied on a limited number of sources \cite{newsroom}.

With the idea of creating extremely short summaries that consist of a single sentence, favoring abstractive summarization strategies over extractive ones to encourage the development of abstractive summarization models, Narayan et al. introduced the Extreme Summarization (XSum) Dataset that was created by gathering articles published in the British Broadcasting Corporation (BBC) website, where each article was paired with a pre-written introductory sentence crafted by the article’s author, who argues that it should succinctly address the question ‘What is the article about?’ in one sentence by utilizing information from different sections of the article, and incorporating techniques of rephrasing, fusion and drawing inferences, unlike headlines, which are designed to catch the reader’s attention \cite{xsum}. 
Due to the nature of the summaries being one-sentence long, they exhibit greater conciseness compared to the summaries in the other datasets.

\begin{figure}[t]
    \centering
    \includegraphics[width=0.48\textwidth]{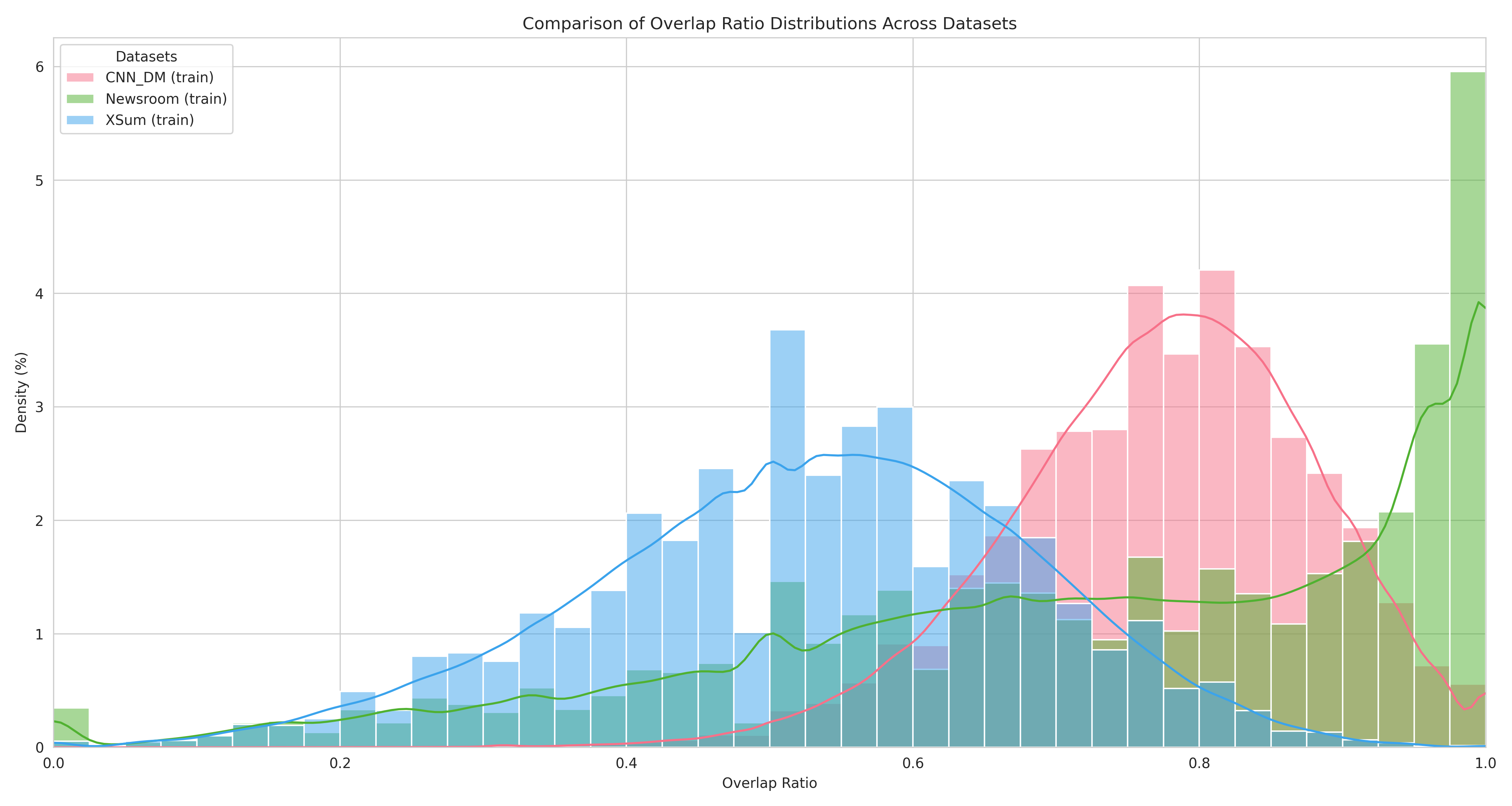}
    \caption{Overlap ratio distributions for the training sets of the three datasets (CNN/DM, Newsroom, XSum), visualized as normalized histograms with overlaid Kernel Density Estimate (KDE) curves. The x-axis represents the overlap ratio, while the y-axis indicates the percentage density, highlighting differences in overlap characteristics among the datasets.}
    \label{fig:overlap_fig}
\end{figure}

In order to gain a deeper understanding of the properties of these datasets, we studied and analyzed the datasets by utilizing visualizations, computing statistical measures such as overlap ratio, and viewing data point examples. 
Based on our comprehensive analysis, we observed the following:
\begin{itemize}
    \item Invalid data points: A small subset of data points was identified where the summary length (in words) exceeded the original text of the article. These instances appear to be incorrect entries, as information presented in the summary does not belong to those mentioned in the corresponding reference article or is not even mentioned in the article text. This situation was encountered in all datasets. This problem was previously mentioned by Chen et al. \cite{chen16}.
    \item Data Quality Issues: Some data points demonstrated various defects and mistakes. Some summaries are totally empty or consist of special characters solely. In addition, certain article texts contain placeholders \texttt{([...])} indicating missing text, which introduces varying degrees of complexity in matching summary information with non-existent article content, thereby posing challenges for effective summarization evaluation. The observed inconsistencies may have arisen from the followed web-based data collection approach and insufficient filtering procedures. These issues were especially encountered in Newsroom dataset.
    \item Different Approaches: Differences in summarization styles across datasets were noticed. The CNN/DM and Newsroom datasets exhibit a bias towards extractive summarization. This is also evident from the high average overlap ratio (approximately 77\% for CNN/DM for example) in the training set, indicating that a significant portion (77\%) of summary words is directly derived from the corresponding article texts. (See Figure \ref{fig:overlap_fig}) In Contrast, the XSum dataset demonstrates a higher degree of abstractive summarization, where the summaries are less reliant on direct extraction from the articles and more on generating entirely novel sentences by paraphrasing the main ideas of the news articles.
\end{itemize}

\subsection{Experimental design}\label{subsec:exp_design}

In this study, we focused exclusively on inferring LMs, employing zero-shot and few-shot in-context learning to assess the performance of various LMs on the news summarization task.
We conducted our experiments in the Google Colab Environment equipped with a high-performance GPU, specifically NVIDIA A100 with 40GB VRAM, which is necessary to utilize the language models.

In the zero-shot setting, the models were assessed on their ability to produce accurate summaries without any helpful context or examples. 
Conversely, in the few-shot setting, a limited number of examples were provided to give minimal guidance, aiming to help the models grasp the nuances of news summarization more effectively, thereby generating more precise and relevant summaries.

The primary reason for not conducting model fine-tuning was the poor quality of the previously examined datasets’ summaries. 
Our analysis indicated that the gold summaries provided were of low quality, and fine-tuning models with such data could lead not only to suboptimal results but might also degrade the models’ summarization abilities. 
The LMs, which inherently have good capabilities in summarization and other NLP tasks, could be negatively affected by the inferior data quality. 
Additionally, resource limitations were a significant challenge in this study. 
Fine-tuning requires considerable computational resources and time, yet we were restricted to the Google Colab's infrastructure, which provides finite compute units upon subscription, which are consumed based on the usage of computational resources. 
Although the initial amount was insufficient and we needed to purchase more and more units to conduct all inferences, the extensive resource requirements for evaluating multiple language models across three datasets in two different settings made fine-tuning both unfeasible and cost-prohibitive.
Moreover, to ensure a clear and focused scope, we decided to concentrate on zero-shot and few-shot settings. 
These approaches also provide valuable insights into the models’ generalization capabilities without the need for additional data.

All the aforementioned reasons apply to public models. 
However, for the tested private models (e.g., GPT-4 or Google Gemini Pro 1.5), fine-tuning was not even an option as this feature was unavailable at the time we conducted our experiments.

\subsection{Experimental settings}\label{subsec:exp_settings}

\subsubsection{Data Integrity}

To maintain data integrity and to ensure robustness and consistency in our work, we implemented a data cleaning procedure to identify and remove invalid data points from the training, validation, and test sets.

\subsubsection{Few-Shots/Demonstrations}

In the few-shot setting, given the constraint of varying context window lengths across different models, it was crucial to include the instructions, the article to be summarized, and the few examples within the prompt, ensuring we did not exceed each model’s context window. 
The distinct tokenization process of each model further complicated this task, making it challenging to provide multiple examples without exceeding the context limit, particularly with longer articles. 
To address this, we decided to include only three examples. 
These examples were manually selected for their critical importance to the experiment, ensuring they were of high quality and represented a variety of genres and topics. 
In some cases, we removed extra, misleading information within the gold summary of those demonstrations to avoid negatively impacting the models’ performance. 
By carefully choosing the shortest articles possible, we aimed to stay within the context window constraints while still providing effective guidance to the models.

\subsubsection{Sampling}

To evaluate the models, we selected a substantial sample of 1000 examples from the test set of each dataset as evaluation examples, respectively. 
This approach contrasts with other works that sampled only 25 to 100 examples as evaluation examples. 
We believe, evaluating on a larger sample size of 1000 examples
\begin{itemize}
    \item guarantees that the results are more statistically robust and reliable, as it reduces the impact of outliers and the variance in performance metrics, leading to more confident conclusions about the model’s capabilities and limitations,
    \item secures that a wider variety of news articles is captured, which helps in assessing the model’s ability to handle different topics, styles, and complexities in news summarization,
    \item provides a better understanding of the model’s generalization capabilities, confirming that it performs well not just on a small, potentially biased subset but across a broader spectrum of real-world scenarios, and
    \item acts as a more rigorous benchmark for future studies, setting a higher standard for model evaluation in the field of news summarization.
\end{itemize}

In contrast to other works (e.g., \cite{zhang23}) which sampled evaluation examples from the validation set and few-shot examples from the training set, we chose to sample the evaluation examples from the test set and the few-shot examples from the validation set.
This approach aligns with potential future research that may involve fine-tuning, where results must be reported on the test set. 
By doing so, we ensure that the evaluation remains consistent and relevant for future comparisons, providing a more reliable benchmark for assessing the impact of fine-tuning on model performance.

\subsubsection{Prompt Design}

Several prompting techniques and strategies should be taken into account while designing effective prompts for In-Context Learning experiments, since these prompts play a pivotal role in determining the models’ success by directing their interactions and outputs. 
This requires not only a deep understanding of the model’s strengths and weaknesses but also domain expertise and a structured method to customize prompts based on each use case.
For instance, when using large models, Zero-shot Learning –where the model performs tasks without any prior examples– often yields satisfactory results due to their advanced capabilities. 
However, smaller models tend to struggle in such scenarios, where employing advanced prompt engineering techniques becomes essential \cite{zhao23, minaee24}. 
Therefore, we ensured through our designed prompts that all LMs comprehended the task requirements regardless of their sizes and complexities to be able to obtain the desired response. The strategies used include
\begin{itemize}
    \item adopting specific role to guide the model’s behavior and to shape the tone and style of the output,
    \item clearly specifying the task to avoid ambiguity and help LMs to understand the scope and constraints,
    \item breaking down the task into multiple steps for better understanding, employing a method akin to the Chain of Thought (CoT) technique which guides LMs through essential reasoning steps, thereby making their implicit processes explicit, and
    \item providing concise and clear instructions, and delivering the article text as input to be summarized by the model.
\end{itemize}

By incorporating these strategies, we aimed to convey our expectations to the LMs clearly and concisely. In \ref{appendix:prompt_template} we present one of our designed prompts.

\subsection{Model Selection}\label{subsec:models}

The selection of models for our research was guided by several key criteria, which can be broadly categorized into constraints for large models and considerations for smaller models.
For large models, our selection was based on their well-known high performance and dominance in various LLM leaderboards. 
Specifically, we included the private OpenAI GPTs and Google Gemini models, recognizing their established performance in the NLP field and demonstrated effectiveness in similar tasks, as well as their widespread use in daily applications such as ChatGPT and Google Gemini (formerly Bard).
For smaller models, we considered the public LMs published on Hugging Face platform (Until May 2024), primarily focusing on the context window length and model size. 

We constrained our model choices to those with a context window length of at least 4096 tokens to be able to fit both prompt and demonstrations in few-shot setting in worst case.
Additionally, due to the computational resources available through the Google Colab Pro+ paid plan, we restricted our selection to models with a maximum parameter count of approximately 11 billion. 
This limitation was necessary to fit the models within the available GPU memory without requiring any type of quantization, which may lead to performance degradation. 
Furthermore, we prioritized the popular models due to their community support.
Based on the aforementioned decisions, we considered benchmarking 20 distinct LMs in both zero-shot and three-shot settings. 
We list the LMs together with their details in table \ref{tab:exp_llms}.

\begin{table}[t]
    \scriptsize
    \centering
    \begin{tabular}{ccccc}
    \hline
    \multirow{2}{*}{Model Name} & \multirow{2}{*}{Creator} & \multirow{2}{*}{\#Parameters} & Context & \multirow{2}{*}{Public} \\
     & & & Window & \\
    \hline
    Gemini-1.5-Pro-0409 \tiny{\cite{gemini}} & Google & - & 128K & \ding{55} \\
    Gemma-2B \tiny{\cite{gemma}} & Google & 2B & 8K & \ding{51} \\
    Gemma-7B \tiny{\cite{gemma}} & Google & 7B & 8K & \ding{51} \\
    GPT-3.5-Turbo-0613 \tiny{\cite{gpt35}} & OpenAI & - & 4K & \ding{55} \\
    GPT-4-0125-preview \tiny{\cite{gpt4}} & OpenAI & - & 8K & \ding{55} \\
    Llama-2-7b-hf \tiny{\cite{llama2}} & Meta & 7B & 4K & \ding{51} \\
    Meta-Llama-3-8B \tiny{\cite{llama3}} & Meta & 8B & 8K & \ding{51} \\
    Meta-Llama-3-8B-Instruct \tiny{\cite{llama3}} & Meta & 8B & 8K & \ding{51} \\
    Mistral-7B-v0.1 \tiny{\cite{mistral}} & Mistral AI & 7B & 4K & \ding{51} \\
    Mistral-7B-Instruct-v0.1 \tiny{\cite{mistral}} & Mistral AI & 7B & 4K & \ding{51} \\
    Phi-3-Mini-4K-Instruct \tiny{\cite{phi3}} & Microsoft & 3.8B & 4K & \ding{51} \\
    Qwen1.5-0.5B \tiny{\cite{qwen}} & Alibaba Cloud & 620M & 32K & \ding{51} \\
    Qwen1.5-1.8B \tiny{\cite{qwen}} & Alibaba Cloud & 1.8B & 32K & \ding{51} \\
    Qwen1.5-4B \tiny{\cite{qwen}} & Alibaba Cloud & 4B & 32K & \ding{51} \\
    Qwen1.5-7B \tiny{\cite{qwen}} & Alibaba Cloud & 7B & 32K & \ding{51} \\
    SOLAR-10.7B-v1.0 \tiny{\cite{solar}} & Upstage & 10.7B & 4K & \ding{51} \\
    SOLAR-10.7B-Instruct-v1.0 \tiny{\cite{solar}} & Upstage & 10.7B & 4K & \ding{51} \\
    Yi-6B \tiny{\cite{yi}} & 01.AI & 6B & 4K & \ding{51} \\
    Yi-9B \tiny{\cite{yi}} & 01.AI & 9B & 4K & \ding{51} \\
    Zephyr-7B-Beta \tiny{\cite{zephyr}} & Hugging Face & 7B & 4K & \ding{51} \\
    \hline
    \end{tabular}
	\caption[List of the selected language models.]{List of the selected language models.}
	\label{tab:exp_llms}
\end{table}

It is important to note that the generation settings (e.g., Temperature, Top-p, etc.) were not altered for either large or small models and kept at their default settings in order to maintain consistency and ensure a fair comparison across different models. 
By using the default settings, we acknowledge that the performance may not be optimal for our specific task, potentially leading to suboptimal results. 
However, our objective was to evaluate the models as they are typically deployed, thereby offering an unbiased assessment of their original performance. 
This is important because different LMs may respond variably to tuned generation settings.

\subsection{Postprocessing}\label{subsec:postprocessing}

During the inferring of private LLMs, we encountered a specific issue that some examples were blocked by the APIs (OpenAI API and Google GenerativeAI API) due to triggering content filters (such as violence, sexual content, self-harm, and hate speech).
Consequently, no completions were generated for these examples, as the content or topics of the news articles activated the filters. 

To ensure a fair and unbiased comparison across the LMs involved in this study, we include only the non-blocked examples in the evaluation process. 
This decision was based on several considerations; firstly, by concentrating on the non-blocked examples, we ensure that all models are assessed using the same set of inputs, thereby eliminating any potential bias introduced by content filtering mechanisms. 
This approach allows us to focus on the models’ core summarization capabilities without the confounding factor of content moderation. 
Secondly, this method prevents penalizing models for complying with content safety guidelines, which is a crucial aspect of their deployment in real-world applications. 
Additionally, it is possible that one or more of the public models are also censored or equipped with similar filters behind the scenes and may likewise generate no completions for those examples, as observed with the private models. 
Thus, we excluded the blocked examples, resulting in a final dataset comprising 827 examples from CNN/DM, 923 examples from Newsroom, and 938 examples from XSum, out of the original 1000 in each case. 
The evaluation of the LMs will be based on these specific subsets.

Another problem faced was that some completions included hallucinated responses. 
For example, while generating the completion, the model would begin to summarize the provided news article but then deviate by producing irrelevant content, such as code snippets, solutions to unmentioned problems, or awkward questions. 
These hallucinations introduce noise into the evaluation process and detract from the models’ primary task of content generation. 
To address this issue, we implemented a basic cleaning procedure to remove such irrelevant completions by identifying known specific patterns, such as (\texttt{\dots\ \textbackslash n\textbackslash n \#\#\# Instructions: HALLUCINATED\textunderscore TEXT}) or the use of triple backticks (\texttt{\dots\ \textasciigrave \textasciigrave \textasciigrave python CODE\textunderscore SNIPPET \textasciigrave \textasciigrave \textasciigrave}) for code snippets. 
However, given the complexity of natural language, it is impossible to anticipate all possible patterns. 
Therefore, the implemented cleaning procedure aims to minimize hallucinations in completions to the greatest extent possible. 
By eliminating these irrelevant continuations, we reduce the variability that could otherwise skew the evaluation outcomes.

\subsection{Evaluation Framework}\label{subsec:eval_framework}

In this study, we implemented an evaluation framework covering automatic, human, and AI-based evaluations methodologies. 

\subsubsection{Automatic Evaluation}

Since manual evaluation of generated summaries on the sampled testing sets is time-consuming, numerous automatic evaluation metrics have been proposed. These metrics typically involve comparing the generated summaries to gold reference summaries. Specifically, our study utilized ROUGE \cite{rouge}, METEOR \cite{meteor}, and BERTScore \cite{bertscore}.

The ROUGE (Recall-Oriented Understudy for Gisting Evaluation) metric measures the quality of candidate (AI-generated) summaries based on the lexical overlap lexical overlap through different approaches (e.g., ROUGE-L examines the Longest Common Subsequence (LCS) between the candidate and reference summaries, capturing sentence-level structural similarities without a specific n-gram length). 
Despite its extensive use, its straightforward implementation and efficiency in measuring lexical overlap, ROUGE has its own limitations, such as its dependence on exact token matches, which means it does not account for synonymous phrases or the semantic meaning of words. 

To address these limitations, our study employs METEOR (Metric for Evaluation of Translation with Explicit ORdering), which incorporates a more sophisticated approach by considering word order, and semantic similarity beyond exact word matches including stems, synonyms, and paraphrastic relationships in its evaluation process. 

However, its reliance on language-specific resources for synonym and paraphrase matching, and the complexity of its calculation led us to explore BERTScore, a significant metric that leverages contextual embeddings from pre-trained language models to address many of the previous metrics' limitations by capturing more nuanced semantic connections between tokens, moving beyond the constraints of exact word matches or predefined synonym sets.
Unlike traditional metrics that struggle with semantic equivalence, unfairly penalizing valid paraphrases, BERTScore can more accurately assess summaries that convey identical meanings using different terminology, thereby correlating better with human judgments. 
Nevertheless, BERTScore faces its own challenges, including increased computational demands and potential biases inherent in the pre-trained language models used. In our study, we specifically utilized the \texttt{"roberta-large"} language model based on the implementation of the \texttt{bert-score} python package. We report F1-scores for these metrics in the results section.

\subsubsection{Human Evaluation Protocol}

Given the large number of summaries generated for each news article, the evaluation task was quite challenging. 
To manage this, we relied on volunteer evaluators who were willing to handle the substantial workload. 
Each evaluator was assigned 6 articles, with 2 articles from each dataset but generated under different settings.
This approach ensured that each evaluator assessed a total of 120 summaries (20 summaries per article × 2 settings/articles × 3 datasets). 
This method was chosen because recruiting a large number of evaluators to handle smaller pieces of work would have required significant coordination and management, increasing the complexity and cost. 
By assigning a substantial number of summaries to each evaluator we ensured that they gained a comprehensive understanding of the range and quality of summaries produced by the models. 
This holistic view allowed evaluators to better distinguish differences in quality that might have been missed if they were only reviewing a smaller subset. 
We understand that the task was overwhelming, and we extend our sincere gratitude to our evaluators for their dedication and hard work.
We instructed five different evaluators to first read the original article carefully before evaluating its summaries. The evaluators were asked to score each summary based on three key aspects:
\begin{itemize}
    \item Relevance: The summary captures the most important information covered in the original article.
    \item Factual Faithfulness: The summary accurately represents the information from the original article without introducing any errors, inconsistencies, or unfaithful details.
    \item Coherence: The summary is logically well-structured and easy to follow. 
\end{itemize}

We used a Likert scale for scoring, ranging from 1 (worst) to 5 (best).

\subsubsection{AI-based Evaluation Protocol}

Using one powerful LM to evaluate others —commonly referred to as LLM-as-a-Judge concept provides a unique way to measure their effectiveness. 
This method not only supports human evaluation but also offers a cost-effective assessment process that could be dynamically adjusted to meet growing evaluation demands.
We followed the same evaluation protocol used for human assessments but applied it to a strong LLM. 
We chose Claude 3 Sonnet, introduced by Anthropic \cite{claude3}, as a judge for this task due to its reputation for exceptional performance across a wide range of tasks and its high ranking on leaderboards. 
Importantly, Claude 3 Sonnet is not related to any of the models in our study, such as GPTs and Gemini, ensuring an unbiased evaluation. 
We used the same three criteria for evaluation. 
To guide the judge LLM, we created a detailed prompt. 
This prompt explained the task, provided the original article text and the candidate summary, and asked the judge LLM to score each criterion following a specific structure.
The judge LLM was asked to assess summaries of five different articles per experiment (20 summaries per article × 5 articles × 3 datasets × 2 settings), resulting in 600 assessments in total.

\section{Experimental results and discussion}\label{sec:results}

\subsection{Zero-shot Learning Results}

In this section, we present the results of our zero-shot learning experiments aimed at evaluating the performance of the LMs considered in this study for the news summarization task. 

\subsubsection{Zero-shot Learning on CNN/DM dataset}

GPT-3.5-Turbo obtained the highest scores in automated metrics on the CNN/DM dataset, with the exception of ROUGE, where Yi-9B had the greatest score. This suggests that while Yi-9B is effective at preserving lexical content and structure, it may be less capable of maintaining semantic coherence and fluency in generated summaries. 
Meanwhile Mistral-Instruct-v0.1 demonstrated the lowest performance across automatic metrics. We hypothesize that this is due to the model’s tendency to truncate the summary generation process too early, leading to summaries that consist of only one or two words. 

In this experiment, we can see that judge LLM preferred some models, such as SOLAR-Instruct-v1.0, Gemma-7B, and Yi-9B, which were not as well regarded by human assessors. Nonetheless, the remaining results were mostly comparable. Humans evaluators and the judge LLM both confirmed on the high performance of Qwen1.5-7B and Llama-3-Instruct regarding relevancy. Furthermore, both concur that Gemma-7B, Qwen1.5-7B, and the Llama-3 family produced relatively faithful summaries on the CNN/DM dataset. Finally, Llama-3-Instruct and Qwen1.5-7B were recognized as the superior models in structuring their summaries. 

Among small models, Gemma-7B, Llama-3 models, Qwen1.5-7B, SOLAR-Instruct-v1.0 and Yi-9B perform particularly well in summarizing news articles into highlights, consistent with the CNN/DM dataset’s characteristics. 

\begin{table*}[t]
	\scriptsize
    \centering
    \begin{tabular}{c|ccc|ccc|ccc}
    \hline
    \multirow{2}{*}{Model Name} & \multicolumn{3}{c}{Automatic Evaluation} & \multicolumn{3}{|c}{Human Evaluation} & \multicolumn{3}{|c}{LLM-as-a-Judge}\\
    & ROUGE-L & BERTScore & METEOR & Relevance & Faithfulness & Coherence & Relevance & Faithfulness & Coherence \\
    \hline
    Gemini-1.5-Pro          & 0.189 & 0.866 & 0.3358 & 4.6 & 5.0 & 4.8 & 5.0 & 5.0 & 5.0 \\
    GPT-3.5-Turbo           & 0.2077 & \textbf{0.8764} & \textbf{0.3613} & 4.4 & 4.8 & 4.8 & 5.0 & 5.0 & 4.8 \\
    GPT-4                   & 0.1643 & 0.8674 & 0.3399 & 4.8 & 5.0 & 4.8 & 4.8 & 5.0 & 5.0 \\
    \hline
    Gemma-2B                & 0.1795 & 0.8542 & 0.1987 & 2.0 & 2.4 & 3.0 & 3.6 & 3.4 & 3.0 \\
    Gemma-7B                & 0.1927 & 0.8578 & 0.2242 & 3.6 & 4.0 & 3.4 & 4.0 & 5.0 & 4.6 \\
    Llama-2-hf              & 0.1653 & 0.8399 & 0.2151 & 3.8 & 4.0 & 4.0 & 3.0 & 3.6 & 3.2 \\
    Llama-3                 & 0.1828 & 0.8584 & 0.2556 & 3.2 & 4.0 & 3.4 & 4.8 & 4.8 & 4.2 \\
    Llama-3-Instruct        & 0.1675 & 0.8495 & 0.301 & 3.8 & 4.0 & 4.2 & 4.8 & 4.6 & 4.0 \\
    Mistral-v0.1            & 0.1698 & 0.8534 & 0.2773 & 3.2 & 2.6 & 3.2 & 4.0 & 3.6 & 3.6 \\
    Mistral-Instruct-v0.1   & 0.1344 & 0.8381 & 0.1587 & 3.4 & 2.6 & 2.6 & 3.0 & 3.2 & 3.4 \\
    Phi-3-Mini-Instruct     & 0.1593 & 0.8523 & 0.2604 & 3.8 & 2.6 & 3.0 & 3.8 & 3.8 & 3.4 \\
    Qwen1.5-0.5B            & 0.1608 & 0.8534 & 0.279 & 2.8 & 1.8 & 2.2 & 3.2 & 2.6 & 2.8 \\
    Qwen1.5-1.8B            & 0.1617 & 0.8502 & 0.268 & 3.4 & 3.2 & 3.4 & 3.4 & 3.2 & 3.2 \\
    Qwen1.5-4B              & 0.1450 & 0.83 & 0.2503 & 3.4 & 3.0 & 3.2 & 3.6 & 4.0 & 3.8 \\
    Qwen1.5-7B              & 0.1735 & 0.8575 & 0.2823 & 4.2 & 4.4 & 4.0 & 4.4 & 4.2 & 4.2 \\
    SOLAR-v1.0              & 0.1534 & 0.855 & 0.2522 & 3.4 & 3.8 & 4.0 & 4.0 & 3.4 & 3.4 \\
    SOLAR-Instruct-v1.0     & 0.1692 & 0.8594 & 0.2887 & 3.6 & 3.2 & 3.6 & 4.6 & 4.6 & 4.6 \\
    Yi-6B                   & 0.1919 & 0.8539 & 0.2409 & 3.6 & 3.4 & 4.2 & 3.6 & 4.0 & 4.2 \\
    Yi-9B                   & \textbf{0.2112} & 0.8649 & 0.2515 & 3.2 & 3.4 & 3.8 & 4.4 & 4.4 & 4.6 \\
    Zephyr-Beta             & 0.1633 & 0.8573 & 0.2894 & 3.6 & 2.8 & 3.8 & 4.0 & 3.6 & 3.8 \\
    \hline
    \end{tabular}
	\caption[Evaluation results for zero-shot LMs on CNN/DM dataset]{Evaluation results for zero-shot LMs on CNN/DM dataset. The highest values in the automatic evaluation metrics are emphasized in bold. }
	\label{tab:zeroshot_cnndm}
\end{table*}

\subsubsection{Zero-shot Learning on Newsroom dataset}

When analyzing the results for the Newsroom dataset shown in Table \ref{tab:zeroshot_newsroom}, we observe a decline in scores compared to the CNN/DM dataset. We attribute this drop to the greater variety of styles that summaries should adhere to, as articles are obtained from a larger number of sources (38 instead of 2). 
In addition, the lower quality of the gold summaries, which we previously discussed during our dataset analysis, is a contributing factor.

The automated metric scores reveal that GPT-3.5-Turbo demonstrates the highest performance across most evaluation metrics. 
In contrast, Llama-2-hf, Llama-3-Instruct, and SOLAR-v1.0 exhibit the lowest scores, suggesting limited effectiveness in this experiment. 
Furthermore, upon examining the summaries generated by these models, it is evident that Llama-2-hf and Llama-3-Instruct failed to produce summaries, resulting in empty completions. 
Specifically, they generated 190 and 203 empty summaries out of a total of 923 articles, respectively. 
While SOLAR-v1.0 tended to output further prompts and instructions rather than fulfilling the task of summarizing the provided news articles. 
These factors significantly impacted their scores in this experiment.

Clearly, we continue to observe that Yi models excel at generating summaries that include n-grams present in the gold summaries, resulting in the highest ROUGE score. 
Additionally, Qwen1.5-7B and SOLAR-Instruct-v1.0 models still perform well in zero-shot news summarization.
Overall, human evaluators favored just two models in this experiment mostly: Qwen1.5-7B and Zephyr-Beta. 
However, judge LLM determined that the Llama-3 family, Mistral-v0.1, Qwen1.5-4B, and SOLAR-Instruct-v1.0 are notably also capable of delivering strong performance.

\begin{table*}
	\scriptsize
    \centering
    \begin{tabular}{c|ccc|ccc|ccc}
    \hline
    \multirow{2}{*}{Model Name} & \multicolumn{3}{c}{Automatic Evaluation} & \multicolumn{3}{|c}{Human Evaluation} & \multicolumn{3}{|c}{LLM-as-a-Judge}\\
    & ROUGE-L & BERTScore & METEOR & Relevance & Faithfulness & Coherence & Relevance & Faithfulness & Coherence \\
    \hline
    Gemini-1.5-Pro          & 0.1704 & 0.8676 & 0.2658 & 4.4 & 5.0 & 4.8 & 4.8 & 5.0 & 5.0 \\
    GPT-3.5-Turbo           & 0.1987 & \textbf{0.8714} & \textbf{0.2915} & 4.2 & 4.8 & 4.8 & 4.6 & 5.0 & 5.0 \\
    GPT-4                   & 0.1684 & 0.8649 & 0.2704 & 4.6 & 4.8 & 4.8 & 4.6 & 5.0 & 5.0 \\
    \hline
    Gemma-2B                & 0.1538 & 0.8507 & 0.1978 & 3.2 & 3.6 & 3.8 & 3.0 & 4.6 & 3.8 \\
    Gemma-7B                & 0.1572 & 0.814 & 0.226 & 2.6 & 3.0 & 3.0 & 3.8 & 4.2 & 4.4 \\
    Llama-2-hf              & 0.1155 & 0.6605 & 0.1557 & 3.8 & 3.2 & 4.2 & 3.4 & 4.2 & 3.8 \\
    Llama-3                 & 0.1152 & 0.6791 & 0.1853 & 2.4 & 2.4 & 3.0 & 4.2 & 5.0 & 4.8 \\
    Llama-3-Instruct        & 0.101 & 0.8086 & 0.1405 & 1.2 & 1.4 & 2.6 & 4.6 & 5.0 & 5.0 \\
    Mistral-v0.1            & 0.1297 & 0.8449 & 0.2231 & 3.2 & 2.6 & 2.6 & 4.0 & 4.6 & 4.0 \\
    Mistral-Instruct-v0.1   & 0.1497 & 0.8504 & 0.1797 & 2.4 & 3.0 & 3.0 & 3.6 & 4.2 & 3.8 \\
    Phi-3-Mini-Instruct     & 0.1294 & 0.8476 & 0.2092 & 3.4 & 3.2 & 3.6 & 4.0 & 4.0 & 4.8 \\
    Qwen1.5-0.5B            & 0.1342 & 0.8531 & 0.2259 & 3.6 & 2.2 & 3.2 & 3.8 & 4.0 & 4.0 \\
    Qwen1.5-1.8B            & 0.1375 & 0.8538 & 0.2299 & 3.6 & 2.8 & 3.8 & 3.8 & 4.6 & 4.8 \\
    Qwen1.5-4B              & 0.1315 & 0.85 & 0.2268 & 3.4 & 3.0 & 3.2 & 4.0 & 4.6 & 4.8 \\
    Qwen1.5-7B              & 0.1481 & 0.8569 & 0.2376 & 3.8 & 3.0 & 4.6 & 4.2 & 5.0 & 4.8 \\
    SOLAR-v1.0              & 0.1079 & 0.8168 & 0.1571 & 3.2 & 1.8 & 3.0 & 3.6 & 4.4 & 4.4 \\
    SOLAR-Instruct-v1.0     & 0.1415 & 0.8536 & 0.222 & 3.6 & 2.4 & 3.6 & 4.0 & 5.0 & 5.0 \\
    Yi-6B                   & 0.1986 & 0.8573 & 0.2098 & 2.6 & 3.4 & 3.4 & 2.8 & 3.2 & 3.2 \\
    Yi-9B                   & \textbf{0.2022} & 0.8626 & 0.1872 & 2.4 & 3.0 & 3.4 & 2.2 & 2.8 & 4.0 \\
    Zephyr-Beta             & 0.1188 & 0.8478 & 0.2216 & 4.2 & 4.0 & 4.4 & 4.2 & 4.8 & 5.0 \\
    \hline
    \end{tabular}
	\caption[Evaluation results for zero-shot LMs on Newsroom dataset]{Evaluation results for zero-shot LMs on Newsroom dataset. The highest values in the automatic evaluation metrics are emphasized in bold.}
	\label{tab:zeroshot_newsroom}
\end{table*}

\subsubsection{Zero-shot Learning on XSum dataset}

Analogous to the scores on the Newsroom dataset, this dataset also shows a drop in scores, which can be attributed to the challenging task of generating highly succinct, single-sentence summaries.

The automatic evaluation scores for the XSum dataset in Table \ref{tab:zeroshot_xsum} manifest the outstanding performance of the Yi-9B model, which achieved the highest scores across ROUGEL (0.2534), and BERTScore (0.8884) metrics. 
This model outperformed even the very large private models, indicating its superior capability of generating extremely concise summaries consisting of one or at most two sentences. 

Notably, this is the first experiment in which the Gemini-1.5-Pro model outperformed the GPT models, achieving the highest METEOR score (0.2923) among all models.

On the other hand, the Gemma-7B, Llama-2-hf, and Llama-3-Instruct models displayed the lowest scores. These models encountered the same issues observed in the zero-shot newsroom experiment. Both Gemma-7B and Llama-2-hf generated a substantial number of empty summaries, 310 and 225 out of 938 articles, respectively, while the Llama-3-Instruct model produced unrelated prompts and instructions instead of actual summaries.

The judge LLM once again aligned with major human selections. 
Both humans and the judge LLM agreed that Zephyr-Beta, Qwen1.5-7B, Gemma-7B, and Solar are significant models for this experiment. 
Nonetheless, there were certain disagreements. 
Although humans recognized the efficacy of Mistral-v0.1, the judge LLM favored the Llama-3 family, Qwen1.5-1.8B, and Qwen1.5-4B.

\begin{table*}
	\scriptsize
    \centering
    \begin{tabular}{c|ccc|ccc|ccc}
    \hline
    \multirow{2}{*}{Model Name} & \multicolumn{3}{c}{Automatic Evaluation} & \multicolumn{3}{|c}{Human Evaluation} & \multicolumn{3}{|c}{LLM-as-a-Judge}\\
    & ROUGE-L & BERTScore & METEOR & Relevance & Faithfulness & Coherence & Relevance & Faithfulness & Coherence \\
    \hline
    Gemini-1.5-Pro          & 0.2197 & 0.8869 & \textbf{0.2923} & 4.4 & 4.8 & 4.8 & 4.4 & 5.0 & 4.8 \\
    GPT-3.5-Turbo           & 0.1934 & 0.8791 & 0.2617 & 4.4 & 4.6 & 4.4 & 4.6 & 5.0 & 5.0 \\
    GPT-4                   & 0.1644 & 0.8718 & 0.2588 & 4.0 & 4.6 & 4.8 & 4.8 & 5.0 & 5.0 \\
    \hline
    Gemma-2B                & 0.1645 & 0.8694 & 0.1924 & 3.0 & 2.4 & 3.8 & 3.8 & 5.0 & 4.8 \\
    Gemma-7B                & 0.1198 & 0.5835 & 0.1529 & 4.0 & 4.0 & 4.6 & 4.4 & 4.8 & 4.8 \\
    Llama-2-hf              & 0.1096 & 0.6522 & 0.1356 & 1.0 & 1.0 & 1.4 & 3.2 & 4.0 & 4.0 \\
    Llama-3                 & 0.1422 & 0.7916 & 0.2144 & 3.4 & 2.8 & 3.8 & 4.0 & 4.8 & 4.8 \\
    Llama-3-Instruct        & 0.1042 & 0.8126 & 0.1498 & 1.6 & 1.6 & 2.8 & 4.0 & 4.8 & 4.8 \\
    Mistral-v0.1            & 0.1281 & 0.8553 & 0.2175 & 4.2 & 3.6 & 4.2 & 3.8 & 3.4 & 4.2 \\
    Mistral-Instruct-v0.1   & 0.159 & 0.8564 & 0.1993 & 3.0 & 3.0 & 3.4 & 3.4 & 4.0 & 3.8 \\
    Phi-3-Mini-Instruct     & 0.1231 & 0.8523 & 0.1947 & 4.0 & 3.0 & 4.2 & 4.2 & 4.4 & 4.6 \\
    Qwen1.5-0.5B            & 0.1387 & 0.8608 & 0.2028 & 3.4 & 2.6 & 3.6 & 3.8 & 3.8 & 4.4 \\
    Qwen1.5-1.8B            & 0.1322 & 0.8585 & 0.2098 & 3.8 & 3.2 & 3.6 & 4.4 & 5.0 & 5.0 \\
    Qwen1.5-4B              & 0.1492 & 0.8635 & 0.2281 & 4.0 & 3.8 & 4.4 & 4.2 & 4.8 & 4.8 \\
    Qwen1.5-7B              & 0.1629 & 0.8669 & 0.2065 & 4.2 & 3.8 & 4.8 & 4.2 & 4.8 & 4.6 \\
    SOLAR-v1.0              & 0.1435 & 0.8580 & 0.2006 & 4.0 & 3.8 & 4.2 & 4.6 & 5.0 & 5.0 \\
    SOLAR-Instruct-v1.0     & 0.1433 & 0.8597 & 0.2118 & 4.2 & 3.6 & 3.8 & 4.0 & 5.0 & 4.6 \\
    Yi-6B                   & 0.2222 & 0.8809 & 0.2325 & 2.8 & 2.6 & 3.2 & 3.8 & 4.8 & 4.6 \\
    Yi-9B                   & \textbf{0.2534} & \textbf{0.8884} & 0.2649 & 2.2 & 1.8 & 2.4 & 3.6 & 4.6 & 4.8 \\
    Zephyr-Beta             & 0.1349 & 0.8572 & 0.2374 & 4.0 & 4.4 & 4.2 & 4.8 & 5.0 & 5.0 \\
    \hline
    \end{tabular}
	\caption[Evaluation results for zero-shot LMs on XSum dataset]{Evaluation results for zero-shot LMs on XSum dataset. The highest values in the automatic evaluation metrics are emphasized in bold.}
	\label{tab:zeroshot_xsum}
\end{table*}

\subsection{Few-shot Learning Results}

In this section, we detail the results of our few-shot (more specifically three-shot) learning experiments in carefully organized tables, followed by an analysis of the results.

\subsubsection{Three-shot Learning on CNN/DM dataset}

Compared to the scores of zero-shot experiment, the results of three-shot experiment for CNN/DM presented in Table \ref{tab:fewshot_cnndm} ensure that offering few-shot examples did not enhance the performance of the LMs; rather, it led to a decline in their scores. 
This is explained by the low quality of the gold summaries, which likely confused the models and hindered their ability to generate accurate summaries. 
Yet, this statement is not valid for large models as they still perform well despite the low-quality demonstrations.

In this experiment, Gemini-1.5-Pro and GPT-3.5-Turbo emerged as the best-performing models, achieving the highest scores. But Gemma-7B and Mistral-v0.1 displayed the lowest scores, indicating significant underperformance. These models responded to more than half of the articles (580 and 555 out of 827 total articles, respectively) with empty answers. 

Among smaller models, Qwen1.5-4B, Qwen1.5-7B, SOLAR-Instruct-v1.0, Yi-6B and Yi-9B demonstrated robust performance in this experiment. Most notably, the Qwen1.5-0.5B model, despite having millions rather than billions of parameters, performed on par with the aforementioned models and outperformed several other larger models looking at their automated metric scores.

The judge LLM gave a high ranking to the Llama-3 family and SOLAR family, while humans add the Gemma family and the Qwen1.5-7B model to this list.

\begin{table*}
	\scriptsize
    \centering
    \begin{tabular}{c|ccc|ccc|ccc}
    \hline
    \multirow{2}{*}{Model Name} & \multicolumn{3}{c}{Automatic Evaluation} & \multicolumn{3}{|c}{Human Evaluation} & \multicolumn{3}{|c}{LLM-as-a-Judge}\\
    & ROUGE-L & BERTScore & METEOR & Relevance & Faithfulness & Coherence & Relevance & Faithfulness & Coherence \\
    \hline
    Gemini-1.5-Pro          & 0.2343 & \textbf{0.8811} & 0.3485 & 4.4 & 5.0 & 4.8 & 5.0 & 5.0 & 5.0 \\
    GPT-3.5-Turbo           & \textbf{0.2377} & 0.8806 & \textbf{0.3525} & 4.2 & 4.8 & 4.8 & 5.0 & 5.0 & 4.8 \\
    GPT-4                   & 0.1947 & 0.8736 & 0.3484 & 4.8 & 5.0 & 4.6 & 4.8 & 5.0 & 4.8 \\
    \hline
    Gemma-2B                & 0.1653 & 0.8525 & 0.1946 & 3.6 & 4.4 & 4.0 & 2.4 & 2.6 & 3.0 \\
    Gemma-7B                & 0.0558 & 0.2573 & 0.0596 & 3.4 & 4.2 & 4.2 & 3.6 & 3.8 & 3.6 \\
    Llama-2-hf              & 0.1518 & 0.79 & 0.179 & 3.4 & 3.4 & 3.4 & 3.0 & 2.6 & 2.4 \\
    Llama-3                 & 0.1694 & 0.7939 & 0.2252 & 3.8 & 3.8 & 4.2 & 3.8 & 4.6 & 4.4 \\
    Llama-3-Instruct        & 0.1612 & 0.8104 & 0.2432 & 4.0 & 4.0 & 4.0 & 4.0 & 4.0 & 4.0 \\
    Mistral-v0.1            & 0.0522 & 0.2773 & 0.0673 & 3.0 & 3.2 & 3.2 & 3.0 & 2.6 & 2.8 \\
    Mistral-Instruct-v0.1   & 0.1368 & 0.6916 & 0.1577 & 3.4 & 3.6 & 3.8 & 2.2 & 3.0 & 2.8 \\
    Phi-3-Mini-Instruct     & 0.1541 & 0.8436 & 0.2359 & 2.6 & 2.4 & 2.6 & 3.4 & 3.6 & 3.2 \\
    Qwen1.5-0.5B            & 0.1654 & 0.8521 & 0.2723 & 3.0 & 2.0 & 2.8 & 3.4 & 3.2 & 3.0 \\
    Qwen1.5-1.8B            & 0.1769 & 0.8536 & 0.2517 & 3.8 & 3.6 & 3.2 & 2.8 & 2.4 & 2.6 \\
    Qwen1.5-4B              & 0.1741 & 0.8548 & 0.277 & 3.6 & 3.6 & 3.6 & 3.0 & 3.2 & 3.2 \\
    Qwen1.5-7B              & 0.1714 & 0.8537 & 0.2556 & 4.4 & 4.2 & 4.2 & 3.8 & 3.8 & 3.4 \\
    SOLAR-v1.0              & 0.1546 & 0.8389 & 0.1994 & 3.8 & 3.6 & 3.6 & 4.0 & 4.0 & 3.6 \\
    SOLAR-Instruct-v1.0     & 0.1682 & 0.8594 & 0.285 & 4.2 & 4.0 & 4.2 & 4.0 & 3.8 & 3.8 \\
    Yi-6B                   & 0.1883 & 0.8624 & 0.1857 & 4.0 & 3.6 & 3.8 & 3.4 & 3.8 & 3.4 \\
    Yi-9B                   & 0.1804 & 0.8574 & 0.1565 & 4.0 & 3.4 & 4.0 & 3.2 & 3.2 & 4.0 \\
    Zephyr-Beta             & 0.1624 & 0.8435 & 0.2641 & 4.0 & 3.6 & 3.8 & 3.6 & 3.6 & 3.8 \\
    \hline
    \end{tabular}
	\caption[Evaluation results for three-shot LMs on CNN/DM dataset]{Evaluation results for three-shot LMs on CNN/DM dataset. The highest values in the automatic evaluation metrics are emphasized in bold.}
	\label{tab:fewshot_cnndm}
\end{table*}

\subsubsection{Three-shot Learning on Newsroom dataset}

When it came to small models in this experiment, Qwen1.5-7B regularly obtained outstanding scores that were consistent across evaluation methodologies. 
Aside from Qwen1.5-7B, models like Qwen1.5-4B, SOLAR-Instruct-v1.0, and Yi-9B illustrated significant effectiveness in automatic measures, although human assessors and judge LLM disagreed, concluding that Gemma-7B, Llama-3, and Zephyr-Beta performed excellently.

Once again, both Gemma-7B and Mistral-v0.1 continue to have serious challenges in generating summaries, failing to produce summaries for 822 and 589 out of 923 total articles, respectively.

\begin{table*}
	\scriptsize
    \centering
    \begin{tabular}{c|ccc|ccc|ccc}
    \hline
    \multirow{2}{*}{Model Name} & \multicolumn{3}{c}{Automatic Evaluation} & \multicolumn{3}{|c}{Human Evaluation} & \multicolumn{3}{|c}{LLM-as-a-Judge}\\
    & ROUGE-L & BERTScore & METEOR & Relevance & Faithfulness & Coherence & Relevance & Faithfulness & Coherence \\
    \hline
    Gemini-1.5-Pro          & 0.1772 & 0.8695 & 0.2632 & 4.4 & 4.6 & 5.0 & 4.4 & 5.0 & 5.0 \\
    GPT-3.5-Turbo           & \textbf{0.2144} & \textbf{0.8744} & \textbf{0.2892} & 4.4 & 4.8 & 4.8 & 4.0 & 5.0 & 5.0 \\
    GPT-4                   & 0.1917 & 0.8707 & 0.2621 & 4.2 & 4.8 & 4.8 & 4.8 & 5.0 & 5.0 \\
    \hline
    Gemma-2B                & 0.122 & 0.8383 & 0.1623 & 2.8 & 2.4 & 3.4 & 3.2 & 4.2 & 4.4 \\
    Gemma-7B                & 0.0176 & 0.0943 & 0.0207 & 4.0 & 3.6 & 4.4 & 3.4 & 4.6 & 4.6 \\
    Llama-2-hf              & 0.1407 & 0.8042 & 0.1728 & 2.4 & 2.6 & 3.4 & 3.0 & 3.6 & 3.2 \\
    Llama-3                 & 0.1118 & 0.5872 & 0.1457 & 4.2 & 3.8 & 4.4 & 4.0 & 4.6 & 4.8 \\
    Llama-3-Instruct        & 0.1062 & 0.7969 & 0.1583 & 2.0 & 2.2 & 2.2 & 4.4 & 5.0 & 5.0 \\
    Mistral-v0.1            & 0.0543 & 0.3082 & 0.0734 & 4.0 & 4.2 & 4.2 & 3.4 & 3.2 & 3.6 \\
    Mistral-Instruct-v0.1   & 0.1645 & 0.7966 & 0.1708 & 3.2 & 3.8 & 3.8 & 2.6 & 3.6 & 3.2 \\
    Phi-3-Mini-Instruct     & 0.1342 & 0.8417 & 0.1936 & 3.6 & 3.8 & 3.4 & 3.8 & 4.0 & 4.6 \\
    Qwen1.5-0.5B            & 0.1302 & 0.8525 & 0.2208 & 3.4 & 3.2 & 2.4 & 3.6 & 3.4 & 4.6 \\
    Qwen1.5-1.8B            & 0.1388 & 0.8544 & 0.2249 & 3.4 & 2.6 & 3.6 & 3.4 & 3.2 & 4.4 \\
    Qwen1.5-4B              & 0.1367 & 0.8542 & 0.231 & 4.2 & 3.6 & 4.0 & 4.2 & 3.6 & 4.8 \\
    Qwen1.5-7B              & 0.1379 & 0.8544 & 0.2258 & 4.0 & 4.4 & 4.6 & 4.2 & 4.2 & 4.8 \\
    SOLAR-v1.0              & 0.1791 & 0.8572 & 0.1814 & 4.2 & 3.8 & 4.2 & 3.2 & 3.8 & 4.0 \\
    SOLAR-Instruct-v1.0     & 0.1777 & 0.8631 & 0.2358 & 4.0 & 3.6 & 4.2 & 3.4 & 3.8 & 4.8 \\
    Yi-6B                   & 0.1728 & 0.8511 & 0.1919 & 2.8 & 3.0 & 3.0 & 2.8 & 3.8 & 4.0 \\
    Yi-9B                   & 0.1905 & 0.8608 & 0.1798 & 2.0 & 2.2 & 2.8 & 3.4 & 4.2 & 4.8 \\
    Zephyr-Beta             & 0.1279 & 0.8175 & 0.2064 & 4.0 & 3.6 & 4.4 & 4.2 & 3.8 & 4.8 \\
    \hline
    \end{tabular}
	\caption[Evaluation results for three-shot LMs on Newsroom dataset]{Evaluation results for three-shot LMs on Newsroom dataset. The highest values in the automatic evaluation metrics are emphasized in bold.}
	\label{tab:fewshot_newsroom}
\end{table*}

\subsubsection{Three-shot Learning on XSum dataset}

Similarly, as observed in the zero-shot experiment on the same dataset (XSum), the Gemini-1.5-Pro model outperformed the GPT models within the category of large models. 
Excellent small models in this experiment that should be acknowledged are SOLAR-v1.0, Yi-6B, and Yi-9B. 
Furthermore, Yi-9B billion even received the highest ROUGE score among all models.

Nonetheless, humans and the judge LLM hold a different perspective. They recognized the strong performance of numerous models, including Gemma-2B, Llama-3, Qwen1.5-7B, SOLAR-v1.0, Zephyr-Beta, and for the first time, Phi-3-Mini-Instruct.

Unfortunately, it appears that certain models, specifically Gemma-7B, Mistral-v0.1, and Llama-3, continue to experience similar performance issues noted in prior experiments. 
They failed to generate 893, 799, and 222 out of 938 summaries, respectively.

\begin{table*}
	\scriptsize
    \centering
    \begin{tabular}{c|ccc|ccc|ccc}
    \hline
    \multirow{2}{*}{Model Name} & \multicolumn{3}{c}{Automatic Evaluation} & \multicolumn{3}{|c}{Human Evaluation} & \multicolumn{3}{|c}{LLM-as-a-Judge}\\
    & ROUGE-L & BERTScore & METEOR & Relevance & Faithfulness & Coherence & Relevance & Faithfulness & Coherence \\
    \hline
    Gemini-1.5-Pro          & 0.2429 & \textbf{0.8929} & \textbf{0.3005} & 4.4 & 4.6 & 4.8 & 4.8 & 5.0 & 5.0 \\
    GPT-3.5-Turbo           & 0.2159 & 0.8856 & 0.2807 & 4.2 & 4.4 & 4.8 & 4.8 & 5.0 & 5.0 \\
    GPT-4                   & 0.1868 & 0.8773 & 0.27 & 4.2 & 4.6 & 4.8 & 4.6 & 5.0 & 5.0 \\
    \hline
    Gemma-2B                & 0.1464 & 0.8497 & 0.1674 & 3.6 & 4.0 & 4.0 & 3.2 & 5.0 & 4.6 \\
    Gemma-7B                & 0.0123 & 0.0426 & 0.0136 & 3.4 & 3.0 & 4.2 & 3.6 & 4.2 & 4.4 \\
    Llama-2-hf              & 0.1693 & 0.8275 & 0.192 & 3.4 & 3.6 & 3.6 & 3.8 & 4.0 & 4.6 \\
    Llama-3                 & 0.1474 & 0.6694 & 0.1817 & 3.6 & 4.2 & 4.0 & 3.8 & 4.4 & 4.6 \\
    Llama-3-Instruct        & 0.1082 & 0.8298 & 0.1787 & 2.4 & 2.4 & 2.6 & 4.8 & 5.0 & 5.0 \\
    Mistral-v0.1            & 0.0226 & 0.127 & 0.0311 & 4.0 & 4.4 & 4.2 & 3.8 & 3.8 & 4.4 \\
    Mistral-Instruct-v0.1   & 0.1744 & 0.8177 & 0.2038 & 3.6 & 3.8 & 4.6 & 4.2 & 4.4 & 4.6 \\
    Phi-3-Mini-Instruct     & 0.1406 & 0.8594 & 0.2038 & 4.0 & 3.8 & 4.6 & 4.2 & 5.0 & 4.8 \\
    Qwen1.5-0.5B            & 0.132 & 0.859 & 0.2061 & 3.6 & 3.2 & 4.2 & 3.8 & 4.2 & 4.6 \\
    Qwen1.5-1.8B            & 0.1447 & 0.8635 & 0.2075 & 3.6 & 3.0 & 4.0 & 4.0 & 3.8 & 4.4 \\
    Qwen1.5-4B              & 0.1546 & 0.867 & 0.2407 & 4.0 & 3.8 & 4.4 & 4.0 & 4.8 & 5.0 \\
    Qwen1.5-7B              & 0.1755 & 0.8717 & 0.2231 & 4.2 & 4.0 & 4.8 & 4.0 & 5.0 & 4.8 \\
    SOLAR-v1.0              & 0.2334 & 0.8723 & 0.2666 & 4.2 & 4.0 & 4.2 & 4.2 & 5.0 & 5.0 \\
    SOLAR-Instruct-v1.0     & 0.1599 & 0.8663 & 0.228 & 3.8 & 3.8 & 4.2 & 3.8 & 5.0 & 4.6 \\
    Yi-6B                   & 0.2322 & 0.8812 & 0.2515 & 2.6 & 3.2 & 3.8 & 3.8 & 5.0 & 4.8 \\
    Yi-9B                   & \textbf{0.2623} & 0.8915 & 0.2808 & 3.0 & 3.2 & 3.0 & 3.4 & 4.2 & 4.0 \\
    Zephyr-Beta             & 0.1578 & 0.8463 & 0.2397 & 4.2 & 4.2 & 4.6 & 4.4 & 4.0 & 4.8 \\
    \hline
    \end{tabular}
	\caption[Evaluation results for three-shot LMs on XSum dataset]{Evaluation results for three-shot LMs on XSum dataset. The highest values in the automatic evaluation metrics are emphasized in bold. }
	\label{tab:fewshot_xsum}
\end{table*}

\subsection{Discussion}

It appears that the judge LLM (Claude 3 Sonnet) tends to assign scores more generously compared to human evaluators. 
But in general, the evaluator LLM aligns with human results in preferring the summaries generated by the large models. 
This is not surprising due to their advanced design, extensive training, and significantly larger parameter counts, which far exceed those of the public smaller models.

Although GPT-3.5-Turbo achieved higher automatic evaluation scores compared to its successor GPT-4, GPT-4 slightly outperformed its predecessor in the majority of Human and AI-Based evaluation metrics. 
This could be attributed to the nature of the summaries generated by GPT-4 since they are more sophisticated, abstractive, and comparable to those that could be authored by humans.

However, upon analyzing the scores of small models only, we noticed that the promising models could be separated into three different patterns regarding the performance:
\begin{itemize}
    \item Models such as Yi-6B, Yi-9B scored remarkably well on automatic evaluation metrics but received poor ratings from both human evaluators and the judge LLM. This suggests these models might be optimizing for surface-level patterns that automatic metrics can capture, rather than producing summaries that humans find useful or accurate.
    \item In contrast, Gemma-7B, Llama-3 family and Zephyr-Beta models received high praise from human evaluators and the judge LLM despite showing modest automatic metric scores, suggesting they better capture the nuances and coherence —aspects that automated metrics might fail to quantify adequately, however humans value them in summaries.
    \item A third set of models, Qwen1.5-7B and SOLAR-Instruct-v1.0, achieved a more balanced performance profile, showing strong automatic metric scores while maintaining respectable human and LLM-judge ratings.
\end{itemize}

Moreover, we observed from the scores given by the judge LLM that small models struggle to produce well-organized summaries for CNN/DM articles, when compared to the levels of coherence seen in summaries generated for Newsroom and XSum datasets in general. 
The structure of the reference summaries, which are highlights, may be responsible for this variation. 
Despite that, it is important to admit that this observation cannot be generalized, as some models, such as Llama-3-Instruct, excel at generating well-structured summaries even for CNN/DM articles.

Finally, we highlight specific limitations we observed in the quality of the generated summaries, based on our analysis of a considerable subset of summaries generated by different LMs. 
The analysis identifies several key issues:
\begin{itemize}
    \item Early Termination of Text Generation: One notable limitation is the tendency of some models to truncate their text generation prematurely. 
    Despite not exceeding the context window, these models often terminate sequences illogically, resulting in incomplete summaries. 
    Models such as Mistral-Instruct-v0.1, Yi-6B, and Yi-9B frequently exhibit this behavior.
    \item Redundancy and Repetitive Sequences: Another significant challenge is generating redundant or repetitive sequences. 
    Some models produce reasonable initial sequences but subsequently regenerate the same content using different wordings or paraphrases, thereby diminishing the overall quality and conciseness of the summaries.
    Models like Phi-3-Mini-4K-Instruct and Mistral-v0.1 are particularly susceptible to this issue.
    \item Generation of Prompts Instead of Task Completions: Certain models tend to generate prompts instead of completing the given tasks. 
    In the context of news summarization, some models respond to the initial prompt with another prompt, failing to generate the required summary. 
    For instance, Llama-3-Instruct frequently produces candidate summaries that outline how to summarize a news article rather than providing the actual summary itself.
    \item Inappropriate Continuation and Hallucinations: A further issue is observed where models generate successful completions but then continue by writing questions, additional prompts or instructions, telling stories, or adding irrelevant information instead of appropriately concluding the text generation process. 
    This problem is notably evident in models like Mistral-v0.1, Llama-3-Instruct, Phi-3-Mini-4K-Instruct, and occasionally in the Qwen1.5 family.
    \item Non-Text Outputs: Lastly, apart from certain models that entirely fail to generate summaries and provide empty outputs such as Gemma-7B, the Llama-2-hf occasionally generates completions that consist solely of newline characters and/or ASCII symbols without producing any meaningful words.
\end{itemize}

\section{Conclusion and future work}\label{sec:conclusion}

In this work, a detailed benchmarking study of 20 recent language models focusing on small models in the context of news summarization was presented using three different datasets: CNN/Daily Mail, Newsroom, and Extreme Summarization (XSum). 
Comprehensive experiments in both zero-shot and few-shot learning scenarios, coupled with diverse evaluation approaches, have provided significant insights into the current state of LMs in this domain.

Our findings show upon analyzing the scores presented in all experiments that large models show superior performance and outperform the smaller models. 
Nevertheless, it appears that models like Qwen1.5-7B, SOLAR-Instruct-v1.0, Llama-3, and Zephyr-Beta are competitive with these large models, as they consistently achieve high scores across all datasets.

Specifically, in the few-shot setting, adding demonstration examples did not improve model performance and even negatively impacted on the ability in extracting the primary concepts mentioned in the article texts, rather than enhancing the model capabilities, primarily due to the low quality of the gold summaries. 
It is noteworthy that large models proved their resilience in maintaining their performance levels across different dataset styles, even in the few-shot setting.

We observed inconsistencies in performance across different datasets, with some models excelling on one dataset while underperforming on others. 
For example, Gemini-1.5-Pro, and SOLAR-v1.0 performed exceptionally on the XSum dataset, highlighting the potential for dataset-specific strengths among LMs.

Our research has highlighted several areas where further investigation and refinement are needed to further improve the evaluation of LMs in the news summarization task:
\begin{itemize}
    \item One significant factor impacting our scores was the quality of the gold summaries in the datasets we used. 
    As detailed previously, the poor quality of these summaries limited our ability to achieve accurate assessments. 
    Therefore, future work should focus on identifying and utilizing comprehensive datasets with high-quality summaries. 
    Fine-tuning LMs on such datasets could yield more reliable outcomes. 
    Alternatively, domain experts could carefully review and refine summaries generated by high-performance LLMs like GPT-4 or Gemini to create high-quality gold summaries for current datasets.
    \item Another potential improvement involves adding a genre attribute to news datasets. 
    By doing so, we could determine which genres present the most challenges for LMs, allowing for targeted, topic-aware analysis and enhancements. 
    This could lead to more nuanced understandings and better performance across different types of news content. 
    \item Additionally, we used default generation settings of LMs in our experiments. 
    Future work should investigate optimal configurations tailored to the summarization task across different models.
    Encouraging model creators to release best-practice configurations for specific tasks would also be beneficial.
    \item Expanding the scope of evaluation to include multi-document news summarization is another promising direction. 
    Our current work focuses specifically on single-document summarization, but a comprehensive evaluation of LMs on multi-document summarization could provide deeper insights and broader applications.
    \item Lastly, future evaluations should not only consider English but also different languages. 
    A multilingual approach would ensure that the findings are applicable across diverse linguistic contexts, enhancing the global relevance and applicability of the research.
\end{itemize}

In conclusion, this benchmarking study has shed light on the effectiveness of recent language models in the news summarization task, emphasizing the continued dominance of large models while also identifying promising small alternatives. 
As the field of natural language processing continues to evolve rapidly, these insights will serve as a valuable foundation for future developments in LM-based summarization technologies.

\section{CRediT authorship contribution statement}\label{sec:credit}
\textbf{Abdurrahman Odabaşı:} Writing – original draft, Validation, Methodology, Investigation, Visualization, Formal analysis, Conceptualization, Funding acquisition.
\textbf{Göksel Biricik:} Writing – review \& editing, Supervision, Project administration, Conceptualization.

\section{Acknowledgments}\label{sec:acknowledgments}

The authors gratefully acknowledge the generous support provided by DAAD (German Academic Exchange Service), which played an essential role in facilitating the computational resources necessary for this research work. This support was instrumental in enabling the experiments and analyses that form the foundation of this study.


\appendix
\section{Designed Prompt Template}
\label{appendix:prompt_template}

In Table~\ref{tab:prompt_xsum_zero}, we present one of the designed prompts, highlighting the components related
to the utilized techniques and approaches with different colors to clarify which part corresponds to which technique.

\begin{table*}
	\scriptsize
    \centering
    \begin{tabular}{|p{4.7in}|}
        \hline
        \texttt{
            \qquad \highlighttext[RGB]{167,224,246}{As a news editor}, \highlighttext[RGB]{253,233,149}{your task is to summarize a news articles in} \highlighttext[RGB]{255,159,127}{a one-sentence summary,} \highlighttext[RGB]{253,233,149}{\quad similar to how a journalist would condense the long article into one sentence that explains the whole story.} 
            \newline \newline \highlighttext[RGB]{225,167,251}{To accomplish this task, start by carefully reading and analyzing the provided news article. Then, use your understanding to capture the most important information, events, and details from the article. Finally, generate a} \highlighttext[RGB]{255,159,127}{one-sentence} \highlighttext[RGB]{225,167,251}{summary that covers the key aspects mentioned in the article.} 
            \newline \newline \highlighttext[RGB]{166,225,197}{Instructions:} 
            \newline \highlighttext[RGB]{166,225,197}{- Ensure that the} \highlighttext[RGB]{255,159,127}{one-sentence} \highlighttext[RGB]{166,225,197}{summary is clear, coherent, informative, succinct, and maintains the original context of the article.} 
            \newline \highlighttext[RGB]{166,225,197}{- Avoid incorporating redundant, unnecessary, or irrelevant information in the summary.} 
            \newline \newline \highlighttext[RGB]{255,159,174}{Article text:} 
            \newline \highlighttext[RGB]{255,159,174}{\{ARTICLE\textunderscore TEXT\} } 
            \newline \newline One-Sentence Summary:
        }\\
        \hline
        \hline
        \textcolor{mycyan}{$\blacksquare$} Role Adoption 
        \qquad\qquad \textcolor{myyellow}{$\blacksquare$} Task Specification 
        \qquad \textcolor{mymor}{$\blacksquare$} Multi-Step Breakdown 
        \newline \textcolor{mygreen}{$\blacksquare$} Clear Instructions 
        \qquad\hspace{0.07cm} \textcolor{myred}{$\blacksquare$} Providing Input 
        \qquad\hspace{0.23cm} \textcolor{myorange}{$\blacksquare$} Length Constraint \\
        \hline
    \end{tabular}
	\caption[Designed Prompt for zero-shot experiments on XSum dataset]{Designed Prompt for zero-shot experiments on XSum dataset}
	\label{tab:prompt_xsum_zero}
\end{table*}


\end{document}